\documentclass[journal]{IEEEtran}
\usepackage{amsmath,amsfonts,amssymb,mathtools}
\usepackage{graphicx}
\usepackage{array}
\usepackage{booktabs}
\usepackage{multirow}
\usepackage{makecell}
\usepackage{tabularx}
\usepackage[table]{xcolor}
\usepackage{url}
\usepackage{cite}
\usepackage{stfloats}
\usepackage{hyperref}
\title{Visual Token Codec: \\ Unleashing Spatial Redundancy for ViT Feature Coding}

\author{
Donghui Feng, Fengxi Zhang,
Changsheng Gao,~\IEEEmembership{Member,~IEEE},
Wenhan~Yang,~\IEEEmembership{Member,~IEEE},
Qi~Wang, Qunshan~Gu, Hongwei~Hu,
Zhengxue~Cheng,~\IEEEmembership{Member,~IEEE}, 
Li~Song,~\IEEEmembership{Senior Member,~IEEE}

\thanks{Donghui Feng, Fengxi Zhang, Zhengxue Cheng, Li Song are with the School of Electronic Information and Electrical Engineering, Shanghai Jiao Tong University, Shanghai, China (e-mail: \{faymek, zhangfengxi, zxcheng, song\_li\}@sjtu.edu.cn).}
\thanks{Changsheng Gao is with the College of Computing and Data Science, Nanyang Technological University, Singapore (email: changsheng.gao@ntu.edu.sg)}
\thanks{Wenhan Yang is with Pengcheng Laboratory, Shenzhen, Guangdong, China (email: yangwh@pcl.ac.cn)}
\thanks{Qi Wang, Qunshan Gu and Hongwei Hu are with the Ant Group, Hangzhou, China. (e-mail: \{qw.qq, qunshan.gu, Hongwei.huhw\}@antgroup.com).}

}

\begin{document}

\maketitle

\begin{abstract}
Distributed deployment of large vision foundation models often partitions a ViT backbone and exchanges intermediate token features between computing nodes, making efficient feature compression critical under bandwidth and computation constraints. Existing ViT feature codecs typically flatten heterogeneous global and patch tokens into an $L \times C$ pseudo image, causing entropy models to mainly capture sequence-axis dependencies while overlooking the native two-dimensional patch-grid structure. In this paper, we show that ViT patch tokens retain strong local spatial correlations on the original grid.
To exploit this structural prior, we propose the \textbf{Visual Token Codec (VTC)}, a dual-path learned codec that separates global and patch tokens into dedicated coding paths. Global tokens are compressed with a lightweight factorized prior, whereas patch tokens are encoded on the patch-token grid using a spatial--channel context entropy model. To support intermediate-layer compression and practical rate adaptation, VTC further incorporates feature-matching supervision after subsequent ViT blocks and variable-rate modules within a single codec.
Experiments on DINOv2 and SAM3 show that VTC consistently outperforms representative ViT feature coding baselines on classification, segmentation, and detection tasks. At 90\% of uncompressed-feature performance, VTC reduces bitrate by \textbf{15.7$\times$--37.4$\times$} across these tasks. We further provide intermediate-layer rate--utility analyses for practical transmission- and storage-oriented deployment scenarios.

\end{abstract}


\section{Introduction}
\label{sec:intro}

\begin{figure}[t]
\centering
\includegraphics[width=\linewidth]{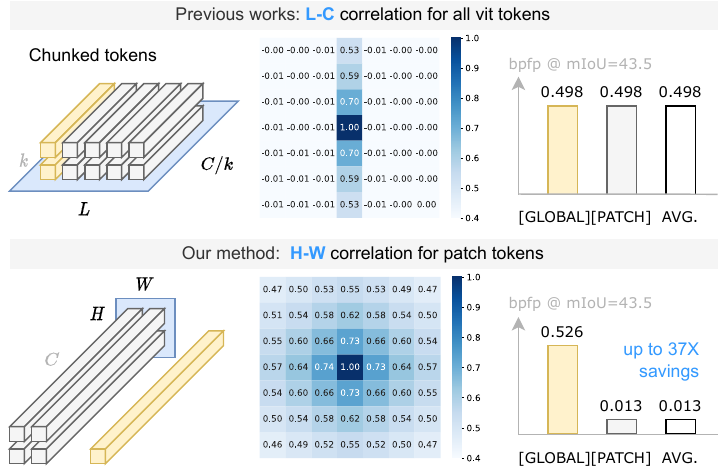}
\caption{Sliding-window 2D Pearson correlations in ViT tokens. Existing methods treat the token sequence (of length $L$ and channel $C$) as a 2D input, merely capturing correlations along the length ($L$) dimension. In contrast, we identify that the dominant patch tokens retain strong spatial correlations ($H \times W$), leading to substantial \textbf{bpfp} (bits per feature point) reduction.}
\label{fig:teaser}
\end{figure}

Recent years have witnessed breakthroughs in large visual foundation models, notably Vision Transformers (ViTs)~\cite{Dosovitskiy2021ViT,Oquab2023DINOv2LR,Tschannen2025SigLIP2M,Ravi2024SAM2S,Carion2025SAM3}, across diverse vision tasks.
However, their high computational demand poses a major challenge for on-device and edge deployment.
A natural solution is distributed deployment, which partitions the ViT backbone across two or more computing nodes~\cite{Ye.2024.OpenFedLLM,Gao.2025.LaMoFC}: a prefix of the model runs on the sender side, and the intermediate features from a selected layer are passed to the receiver side for the remaining computation.
While this design shifts part of the computation away from resource-constrained devices, it introduces a new communication bottleneck: once inference is partitioned, system cost is determined not only by FLOPs, but also by the bandwidth and latency required to exchange high-dimensional activations.
Efficient compression of intermediate ViT features is therefore a practical prerequisite for deploying large ViTs in bandwidth- and computation-constrained environments.

\textbf{ViT feature coding} addresses this bottleneck by formulating intermediate ViT representations as coding targets.
Given a selected compression layer, the ViT prefix transforms the input image into intermediate tokens; a feature encoder maps these tokens into quantized latents and compresses them into a compact bitstream; the receiver decompresses the bitstream, reconstructs the tokens, and feeds them into the remaining ViT layers and frozen downstream heads.
Different from image coding, whose fidelity is measured mainly in the pixel domain, ViT feature coding minimizes bitrate while preserving the utility of reconstructed features for subsequent model computation.
The central objective is therefore rate--distortion optimization in the feature domain, where distortion is reflected by representation fidelity or downstream task performance.

\textbf{Existing ViT feature codecs remain high bitrate, failing to exploit the structural prior of ViT tokens.}
Intermediate ViT features contain a few \textbf{global tokens} (\texttt{[CLS]}, \texttt{[REG]})~\cite{Darcet.2024.ViTRegisters} and many \textbf{patch tokens} arranged on the image patch grid~\cite{Dosovitskiy2021ViT,Oquab2023DINOv2LR}.
However, prior codecs~\cite{Gao.2025.LaMoFC,Gao.2025.DT-UFC,Chen.2026.VQFC} typically flatten these heterogeneous tokens into an $L \times C$ pseudo image for traditional~\cite{Bross2021VTM} or learned~\cite{Balle.2018.Hyperprior18,He.2022.ELIC} image codecs.
As shown by the sliding-window Pearson correlations in Fig.~\ref{fig:teaser}, such a sequence-oriented layout mainly captures dependence along the token-length dimension, while missing the strong local dependence of patch tokens on the original 2D grid.
This mismatch leaves patch-grid redundancy insufficiently modeled and limits rate--distortion performance.

\begin{figure*}[ht]
    \centering
    \includegraphics[width=\linewidth]{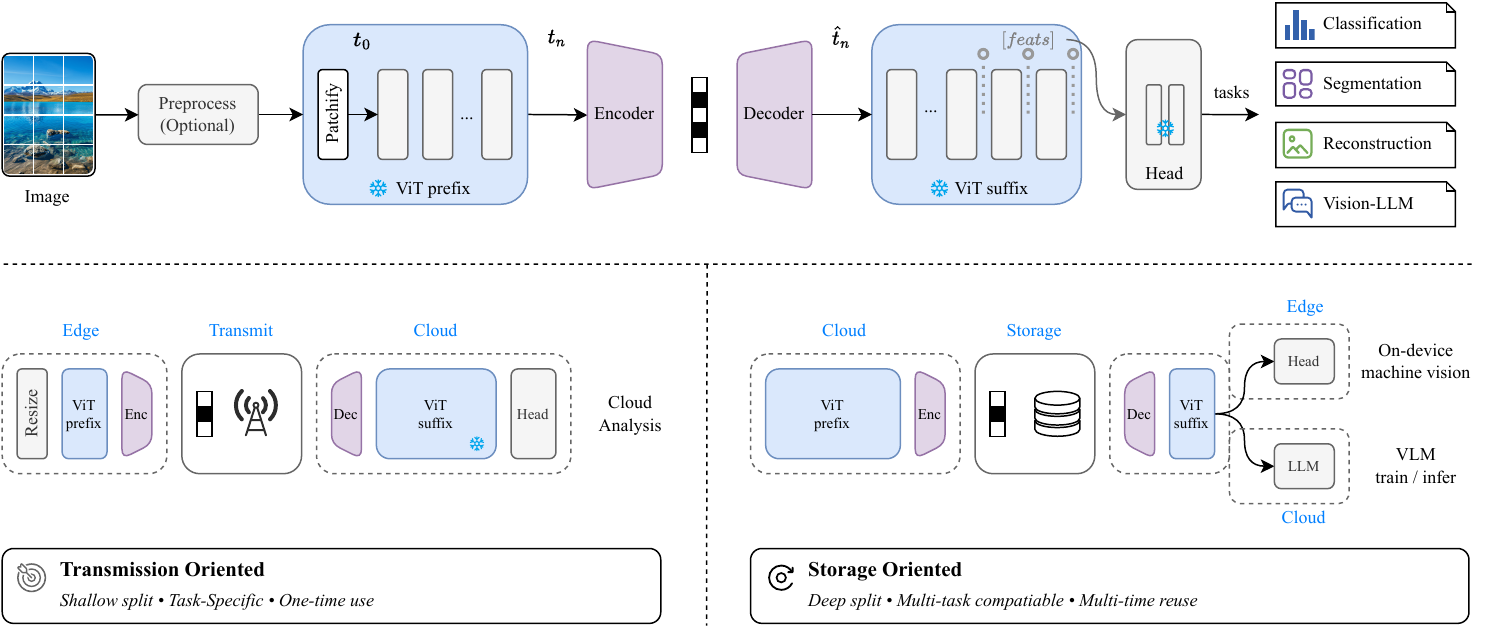}
    \caption{Overview of ViT feature coding and its two deployment paradigms. The upper part illustrates the generic feature coding pipeline, where an input image is processed by a ViT prefix, compressed into a compact bitstream by a feature encoder, transmitted or stored, and then decoded for the remaining ViT layers and frozen downstream heads. The lower-left part shows the transmission-oriented paradigm, which compresses shallow intermediate features for one-time, task-specific edge--cloud inference under bandwidth-constrained links. The lower-right part shows the storage-oriented paradigm, which tends to compress deeper features to generate reusable, multi-task-compatible representations that can be stored once and consumed repeatedly by different downstream tasks.}
    \label{fig:framework}
    \end{figure*}

\textbf{Existing studies also lack a systematic analysis of intermediate-layer feature coding.}
Practical deployments may impose different computation, bandwidth, latency, and reusability requirements, making the preferred compressed layer dependent on the deployment objective.
However, most prior work focuses on compressing final-layer ViT features, while intermediate layers are rarely examined as coding targets.
As illustrated in Fig.~\ref{fig:framework}, \textbf{transmission-oriented feature coding} targets one-time, task-specific edge--cloud inference and often favors compressing a shallow layer to limit on-device computation, whereas \textbf{storage-oriented feature coding} tends to favor compressing a deeper layer to produce reusable representations that can be stored once and consumed by different downstream tasks.
As a result, it remains unclear how to choose the compressed layer that best balances bitrate, on-device computation, and downstream utility.

To address these limitations, we introduce the \textbf{Visual Token Codec (VTC)}, a dual-path framework that routes global and patch tokens through separate entropy models.
Global tokens are compressed with a lightweight factorized prior, while patch tokens are encoded by a Spatial-Channel Context (SCCTX) entropy model~\cite{He.2022.ELIC} on the patch-token grid.
By matching the entropy model to the heterogeneous structure of ViT tokens, VTC explicitly exploits patch-grid spatial priors for higher feature-domain coding efficiency. 
To support intermediate-layer compression, we further introduce a \textbf{feature-matching} loss that aligns deeper features after subsequent ViT blocks, thereby reducing the propagation of quantization-induced distortions.
For practical deployment, VTC integrates \textbf{variable-rate} modules within a single codec.
We instantiate VTC on DINOv2 and SAM3 image encoders to analyze rate--utility trade-offs across compressed layers under both transmission-oriented and storage-oriented scenarios.

The main contributions of this paper are as follows:

\begin{itemize}
\item \textbf{Visual Token Codec}: We propose VTC, a learned codec for ViT intermediate features that separates global and patch tokens into dedicated coding paths. It combines factorized global-token coding, patch-grid context modeling, feature-matching supervision, and variable-rate modules.
\item \textbf{Efficient Feature Coding}: For final-layer feature coding, VTC substantially reduces the bitrate required to preserve downstream performance. At 90\% of uncompressed-feature performance, VTC achieves 15.7$\times$--37.4$\times$ bitrate reduction over LaMoFC VTM, mainly by exploiting the patch-grid structural prior that flattened pseudo-image layouts overlook (Fig.~\ref{fig:teaser}).
\item \textbf{Intermediate-Layer Deployment Study}: Beyond final-layer compression, we provide RD curves for coding different intermediate layers, revealing the importance of the feature-matching supervision layer. We further instantiate lightweight-encoder and lightweight-decoder cases to offer initial bitrate--computation--utility trade-offs for practical ViT feature coding (Fig.~\ref{fig:framework}).
\end{itemize}

The remainder of this paper is organized as follows.
Section~\ref{sec:related} reviews learned image compression and feature coding for CNNs and ViTs.
Section~\ref{sec:method} presents the dual-path VTC framework, feature-matching training, and Variable-Rate Model.
Section~\ref{sec:exp} reports rate--distortion results on DINOv2 and SAM3, together with intermediate-layer analyses and ablation studies.
Section~\ref{sec:conclusion} concludes the paper and discusses limitations and future work.

\section{Related Works}
\label{sec:related}

\subsection{Learning-Based Image Compression}
Over the past decade, learning-based image compression has emerged as a dominant paradigm, often referred to as nonlinear transform coding (NTC~\cite{Balle.2021.NTC}). In NTC, neural-network based nonlinear analysis and synthesis transforms map the image into a latent representation and back, aiming to decorrelate the image signal; an entropy model then captures the latent distribution for efficient coding. Below we briefly review these two core components.

\textbf{Non-linear transform networks:}
Early methods commonly adopted convolutional neural networks (CNNs) with generalized divisive normalization (GDN)~\cite{Balle.2017.Balle17,Balle.2018.Hyperprior18,Minnen.2018.Joint18} in the analysis and synthesis transforms.
Subsequent works enhanced representational capacity by integrating attention mechanisms and residual blocks into variational autoencoder architectures~\cite{Cheng.2020.Cheng20}, or by exploring variants such as invertible networks~\cite{Xie.2021.EIE,Tu.2025.MIN} and wavelet-like transforms~\cite{Ma.2020.EOV}.
Motivated by the success of Transformers in computer vision, recent codecs incorporate architectures such as Swin Transformers~\cite{Zhu.2021.TBTC}, hybrid CNN--Transformer designs~\cite{Liu.2023.TCM-LIC}, and frequency-aware Transformers~\cite{Li.2023.FAT-LIC} into the transform networks.
To mitigate the quadratic cost of self-attention, some studies explore efficient attention alternatives~\cite{Lu.2022.TinyLIC,Qin.2024.MambaVC,Zeng.2025.MambaIC,Wu.2025.CMamba,Chen.2025.CMIC,Feng.2025.LALIC}.
For practical deployment, recent low-complexity works~\cite{Wang.2026.CSLIC,Wang.2023.EVC,Jia.2025.DCVC-RT} further balance rate--distortion performance against encoder/decoder cost.
Overall, learned codecs have trended toward stronger nonlinear transforms that consistently improve decorrelation and rate--distortion performance.

\textbf{Entropy modeling:}
Entropy modeling is crucial for learned image compression because it removes remaining redundancy in latent representations. Balle et al.\ first proposed a factorized prior~\cite{Balle.2017.Balle17} to model marginal distributions, and subsequently introduced a hyperprior~\cite{Balle.2018.Hyperprior18} that uses side information to parameterize a conditional Gaussian distribution. Remaining redundancy is further reduced with autoregressive priors that split latents along spatial or channel dimensions and predict unencoded part from already encoded part, thereby modeling conditional distributions and exploiting spatial and channel correlations. These approaches can be categorized into spatial autoregressive~\cite{Minnen.2018.Joint18,Cheng.2020.Cheng20,He.2021.Checkerboard}, channel autoregressive~\cite{Minnen.2020.Charm}, or combined spatial--channel designs~\cite{He.2022.ELIC,Jiang.2023.MLIC,Jiang.2023.MLIC++}.
Transformer- and context-based entropy models~\cite{Qian.2022.Entroformer,Koyuncu.2022.Contextformera,Koyuncu.2024.EfficientContextformera,Li.2020.LCN} further refine distribution estimation by exploiting broader spatial and channel contexts.

\subsection{Feature Coding for CNNs}
\label{subsec:feature_comp}

The rise of Convolutional Neural Networks (CNNs) in computer vision created a demand for compressing and transmitting intermediate features for machine analysis, giving rise to feature coding as a branch of Video Coding for Machines~\cite{Duan.2020.VCM,Yang.2024.VCM,Li.2024.SIP}. This has led to standardization efforts, with MPEG-FCM~\cite{WG2023Call} being a notable example, where the evaluation protocol jointly considers bitrate and task accuracy. Existing methods typically derive from established image or video codecs and can be grouped by their coding framework.

\textbf{Codec-based packing.}
One common approach adapts traditional video codecs to feature compression. These methods quantize and pack multi-layer feature maps into a pseudo-video frame, which is then encoded by a standard codec such as VTM~\cite{Bross2021VTM}. A representative work by Chen et al.~\cite{Chen.2020.TIP} is widely used as an anchor in MPEG-FCM. Learned video codecs~\cite{Shi.2022.AlphaVC,Xiang.2023.MIMT,Zhang.2025.FLAVC,Jia.2025.DCVC-RT} provide complementary options when feature maps are packed as pseudo-video.

\textbf{Learned feature compression.}
Another line of research designs neural networks to transform or fuse input features into a compact latent representation for entropy coding. Kim et al.~\cite{Kim.2024.TCSVT} propose a framework that fuses multi-scale features, also analyzing task-specific versus reconstruction-oriented training and adopting feature-level MSE for cross-task generality. Liu et al.~\cite{Liu.2023.ICASSP} introduce a coding strategy where compressed lower-scale features help predict higher-scale ones, exploiting inter-scale redundancy. Other works exploit feature-specific properties more directly, such as channel-wise sensitivity for bit allocation~\cite{hu2020sensitivity} and 3D sparse convolutions for low-spatial, high-channel, and sparse feature tensors~\cite{ma2024feature}.

\textbf{Task-aware coding.}
While many learned methods optimize generic feature distortion (e.g., MSE), several works tailor the compression objective to downstream tasks. Choi et al.~\cite{Choi.2021.ICIP} structure the latent space so that a subset of channels supports object detection under a task-specific rate--distortion term, while the full representation enables input reconstruction. Gao et al.~\cite{Gao.2024.DMOFC,Gao.2025.IMOFC} go further by replacing instance-level MSE with discrimination- and identity-level metrics, explicitly optimizing for retrieval and re-identification accuracy. Recent studies further question whether gains come from joint optimization itself or the downstream semantic parser~\cite{gao2025rethinking}, and introduce compressed feature quality assessment to quantify semantic degradation beyond simple feature distances~\cite{gao2025compressed}.

\textbf{Scalable coding.}
Orthogonal to the choice of distortion, scalability organizes the bitstream into layers to serve different quality or task requirements. Scalable image codecs often split the bitstream into a base and an enhancement layer, e.g., a deep-feature base with a texture residual for face images~\cite{Wang.2019.ICIP,yang2021towards}, or semantics-to-signal coding with learned structural representations~\cite{Yan.2021.TIP,Liu.2021.IJCV}. For intermediate features, Chen et al.~\cite{Chen.2024.ICIP} learn scalable multilayer compression across network stages, enabling machine vision at different bit budgets.

\subsection{Feature Coding for Vision Transformers}
\label{subsec:feature_comp_vit}

In the era of vision foundation models, large pretrained Vision Transformers (ViTs)—such as DINOv2~\cite{Oquab2023DINOv2LR}, SigLIP2~\cite{Tschannen2025SigLIP2M}, and SAM2/SAM3~\cite{Ravi2024SAM2S,Carion2025SAM3}—provide robust and generalizable intermediate features that are widely used across diverse vision tasks. Compressing these ViT features is therefore crucial for efficient distributed training and inference, where intermediate representations must be transferred or stored. Recent studies on large-model feature coding further show that modern models produce heterogeneous features with different distributions and compression tolerances, motivating unified benchmarks and codecs beyond conventional CNN feature coding~\cite{Gao.2025.LaMoFC,pang2026towards}.

Early exploration in this area was often task- or model-specific. For instance, an initial study \cite{Duan.2024.CSR} focused solely on compressing the \texttt{[CLS]} token of DINOv2 for image classification.
To establish a common ground for evaluation, comprehensive benchmarks \cite{Gao.2025.LaMoFC,pang2026towards} are introduced, compressing features from diverse models (e.g., DINOv2, LLaMA-3, SD3) across multiple tasks and split-computing settings. Building upon this direction, universal coding frameworks are proposed to improve generalization across feature sources: DT-UFC~\cite{Gao.2025.DT-UFC} employs a learned distribution transformation to align features from different models into a common space, while cross-architecture feature coding~\cite{gao2025cross} further studies distribution alignment between CNN and Transformer features.

Despite these advances, existing approaches predominantly focus on modeling token distributions or feature-value alignment while largely neglecting the spatial organization of ViT tokens. LaMoFC~\cite{Gao.2025.LaMoFC} treats the $L\times C$ token sequence as a pseudo-image and compresses it with a hyperprior, but it does not distinguish between the globally aggregated \texttt{[CLS]} token and patch tokens. DT-UFC~\cite{Gao.2025.DT-UFC} and cross-architecture alignment~\cite{gao2025cross} improve distribution compatibility, yet their coding pipelines still do not explicitly exploit 2D patch-token locality. VQFC~\cite{Chen.2026.VQFC} adopts vector quantization under the assumption of weak spatial redundancy; however, our empirical observations on dense ViT features reveal that strong local correlations still exist in the patch token grid. This oversight makes it challenging to achieve high compression efficiency without compromising task performance, highlighting the need for more effective coding strategies tailored to the structural prior of ViT features.

\section{Proposed Method}
\label{sec:method}

\begin{figure*}[ht]
    \centering
    \includegraphics[width=\textwidth]{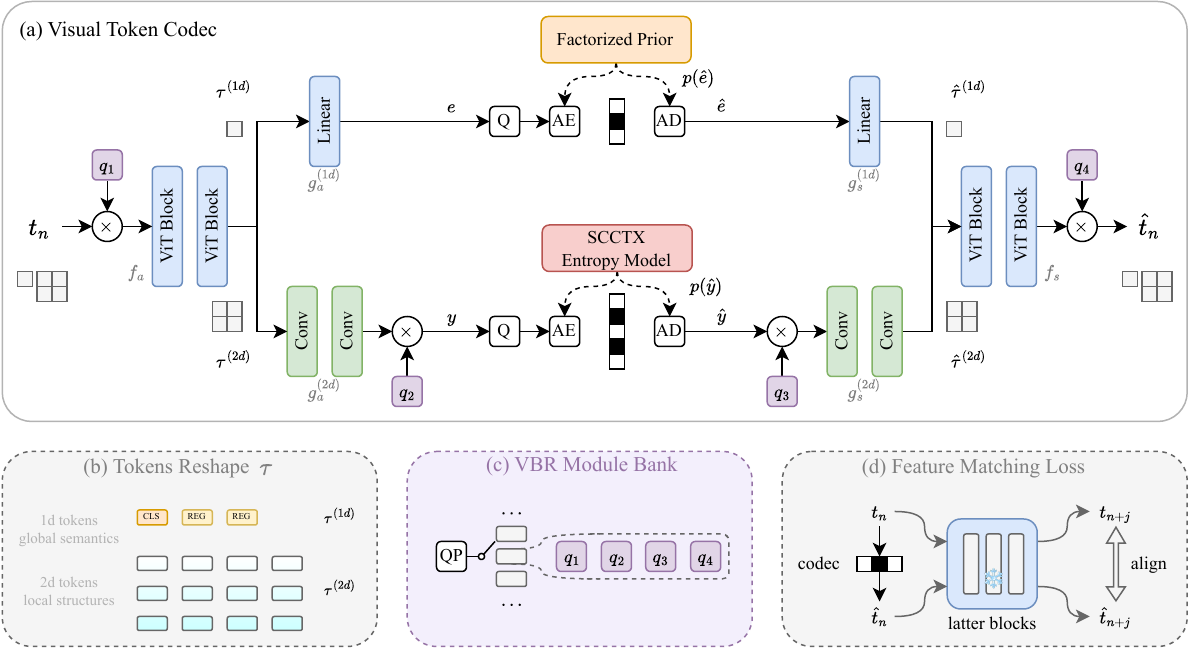}
    \caption{Overview and key components of the proposed Visual Token Codec (VTC).
    (a) Overall architecture. The input feature $\mathbf{t}$ is first transformed into $\boldsymbol{\tau}$ by ViT transform blocks and then compressed through two parallel branches: a global-token path with factorized entropy coding and a patch-token path with spatial--channel context modeling.
    (b) Token Reshape. Global tokens are kept as a short 1D sequence, while patch tokens are restored to their native 2D patch grid to expose strong local spatial redundancy.
    (c) VBR Module Bank. A QP-conditioned module bank modulates the codec features and enables multiple bit rates with a single model.
    (d) Feature Matching Loss. The reconstructed tokens are propagated through subsequent frozen ViT blocks, and the loss is computed on deeper features to reduce representation drift after decoding.
    }
    \label{fig:codec}
\end{figure*}

\subsection{Overall Framework}

As shown in Fig.~\ref{fig:framework}, we partition the ViT backbone into the prefix part and suffix part and insert the codec between them. On the encoder side, the prefix part receives an input image, extracts intermediate features, and passes them to the codec, which compresses the features into a bitstream. On the decoder side, the codec receives the bitstream, reconstructs the features, and then forwards them through the suffix part to extract the features needed for downstream tasks. These features are finally fed into lightweight task heads (conv or linear layer) to perform downstream tasks such as image classification and semantic segmentation.

\subsection{Dual-Path Visual Token Codec}

ViT tokens are heterogeneous: a few \texttt{[CLS]}/\texttt{[REG]} tokens encode global semantics, while the majority are \texttt{[PATCH]} tokens on an $H\times W$ grid with strong local spatial correlation (Fig.~\ref{fig:teaser}).
Compressing both groups with one layout, whether a flattened sequence or a single pseudo-image rasterization, forces one entropy model to adopt priors that fit only one token type. Thus, we propose routing global and patch tokens through separate coding paths.

Given an input image $\mathbf{x}$ and a chosen compressed layer index $n$, the ViT first applies a patch embedding $\mathrm{Emb}(\cdot)$, followed by the first $n$ blocks to get the intermediate feature $\mathbf{t}_n$.
\begin{equation}
\begin{aligned}
\mathbf{t}_0 & = \mathrm{Emb}(\mathbf{x}), \\
\mathbf{t}_n & = \mathrm{Block}_{1:n}(\mathbf{t}_0),
\end{aligned}
\end{equation}

To compress $\mathbf{t}_n$, it is first transformed by $f_a(\cdot)$ into latent $\boldsymbol{\tau}_n$, then passed through the dual-path codec to obtain reconstructed latent $\hat{\boldsymbol{\tau}}_n$, and finally mapped back to the original token space as $\hat{\mathbf{t}}_n$ via $f_s(\cdot)$:
\begin{equation}
\begin{aligned}
\boldsymbol{\tau}_n & = f_a(\mathbf{t}_n), \\
\boldsymbol{\tau}_n & = \bigl[\boldsymbol{\tau}_n^{(1\mathrm{d})}, \boldsymbol{\tau}_n^{(2\mathrm{d})}\bigr], \\
\hat{\mathbf{t}}_n & = f_s(\hat{\boldsymbol{\tau}}_n).
\end{aligned}
\end{equation}
Here, $f_a(\cdot)$ and $f_s(\cdot)$ consist of several learnable ViT blocks that serve as the primary nonlinear transforms. We split $\boldsymbol{\tau}_n$ for dual-path coding: $\boldsymbol{\tau}_n^{(1\mathrm{d})}$ collects the \texttt{[CLS]} and \texttt{[REG]} tokens and is compressed with a lightweight factorized entropy model, while $\boldsymbol{\tau}_n^{(2\mathrm{d})}$ stacks all \texttt{[PATCH]} tokens and is compressed by a Spatial-Channel Context (SCCTX) 2D entropy model proposed in ELIC~\cite{He.2022.ELIC}.
Decoded patch and global branches are merged before $f_s$, so both paths must reconstruct adequately for downstream ViT blocks to approximate uncompressed activations.


\subsection{Global Tokens Coding}
\label{subsec:global_tokens_coding}


To compress the global-token branch $\boldsymbol{\tau}_n^{(1\mathrm{d})}$, an analysis transform $g_a^{(1\mathrm{d})}(\cdot)$ first maps it to a lower‑dimensional latent representation $\mathbf{e}$, which is then quantized to $\hat{\mathbf{e}} = Q(\mathbf{e})$ and entropy‑coded at a rate $R(\hat{\mathbf{e}})$. The decoded tokens are reconstructed by a synthesis transform $g_s^{(1\mathrm{d})}(\cdot)$. The overall process is:
\begin{equation}
    \begin{aligned}
        \mathbf{e}                         & = g_a^{(1\mathrm{d})}(\boldsymbol{\tau}_n^{(1\mathrm{d})}), \\
        \hat{\mathbf{e}}                   & = Q(\mathbf{e}),                                     \\
        \hat{\tau}_n^{(1\mathrm{d})} & = g_s^{(1\mathrm{d})}(\hat{\mathbf{e}}).
    \end{aligned}
\end{equation}

\textbf{Factorized Prior.} We model the distribution of $\hat{\mathbf{e}}$ using a learned factorized prior~\cite{Balle.2018.Hyperprior18}. A factorized prior models each latent element independently with a univariate parametric density. For a scalar latent variable, its cumulative distribution function (CDF) $C$ is constructed as a composition of $K$ learnable transformation layers:
\begin{equation}
    C = f_K \circ f_{K-1} \circ \cdots \circ f_1,
\end{equation}
where each $f_k$ typically consists of linear mapping and monotonic nonlinear activation. 
From the CDF, the probability mass function (PMF) of the quantized latent $\hat{e}$ is obtained by:
\begin{equation}
    p(\hat{e}) = C(\hat{e} + \tfrac{1}{2}) - C(\hat{e} - \tfrac{1}{2}),
\end{equation}
The expected bit‑rate for $\hat{\mathbf{e}}$ is then given by:
\begin{equation}
    R(\hat{\mathbf{e}}) = \mathbb{E}\left[ -\log_2 p(\hat{\mathbf{e}}) \right].
\end{equation}

\subsection{Patch Tokens Coding}

To compress the patch-token branch $\boldsymbol{\tau}_n^{(2\mathrm{d})}$, the analysis transform $g_a^{(2\mathrm{d})}(\cdot)$ first maps it to a lower‑dimensional latent $\mathbf{y}$, which is then quantized and decoded by a synthesis transform $g_s^{(2\mathrm{d})}(\cdot)$:
\begin{equation}
    \begin{aligned}
        \mathbf{y}                         & = g_a^{(2\mathrm{d})}(\boldsymbol{\tau}_n^{(2\mathrm{d})}), \\
        \hat{\mathbf{y}}                   & = Q(\mathbf{y}),                                     \\
        \hat{\mathbf{t}}_n^{(2\mathrm{d})} & = g_s^{(2\mathrm{d})}(\hat{\mathbf{y}}).
    \end{aligned}
\end{equation}

\textbf{SCCTX Entropy Model.} To effectively capture the 2D correlations of $\hat{\mathbf{y}}$, we adopt the existing Spatial-Channel Context (SCCTX) model~\cite{He.2022.ELIC} along with the hyperprior model \cite{Balle.2018.Hyperprior18}.
For the hyperprior, a hyper encoder $h_a(\cdot)$ produces hyper latents that provide side information for entropy modeling, and a hyper decoder $h_s(\cdot)$ maps the quantized hyper latents into a hyperprior context:
\begin{equation}
    \begin{aligned}
        \mathbf{z}                      & = h_a(\mathbf{y}),       \\
        \hat{\mathbf{z}}                & = Q(\mathbf{z}),         \\
        \boldsymbol{\Phi}_{\mathrm{hp}} & = h_s(\hat{\mathbf{z}}),
    \end{aligned}
\end{equation}
where $\hat{\mathbf{z}}$ is entropy‑coded with another factorized prior at rate $R(\hat{\mathbf{z}})$.
The latent $\mathbf{y}$ retains redundancies along both spatial and channel axes.
To exploit these correlations, we partition $\hat{\mathbf{y}}$ into multiple groups and sequentially encode each group conditioned on previously decoded ones.
For the channel dimension, we split the features into $K$ chunks.
The channel context for the $k$-th chunk is computed by a channel‑context network $g_{\mathrm{ch}}^{(k)}(\cdot)$ as:
\begin{equation}
    \boldsymbol{\Phi}_{\mathrm{ch}}^{(k)} = g_{\mathrm{ch}}^{(k)}\left(\hat{\mathbf{y}}^{<k}\right), \quad k=2, \dots, K,
\end{equation}
where $\hat{\mathbf{y}}^{<k} = \{\hat{\mathbf{y}}^{(1)}, \dots, \hat{\mathbf{y}}^{(k-1)}\}$ denotes the set of previously encoded channel chunks.

In the spatial dimension, we adopt a checkerboard partitioning that divides each chunk into anchor and non-anchor positions. Anchor positions are encoded without spatial context, so we set $\boldsymbol{\Phi}_{\mathrm{sp}}^{(k)} = \mathbf{0}$. For non-anchor positions, we compute spatial context from already decoded neighbors using a checkerboard mask $M_{\mathrm{cb}}$ and a spatial‑context network $g_{\mathrm{sp}}^{(k)}$:
\begin{equation}
    \boldsymbol{\Phi}_{\mathrm{sp}}^{(k)} = g_{\mathrm{sp}}^{(k)}\bigl(M_{\mathrm{cb}} \odot \hat{\mathbf{y}}^{(k)}\bigr).
\end{equation}

\begin{figure}[t!]
    \centering
    \includegraphics[width=\linewidth]{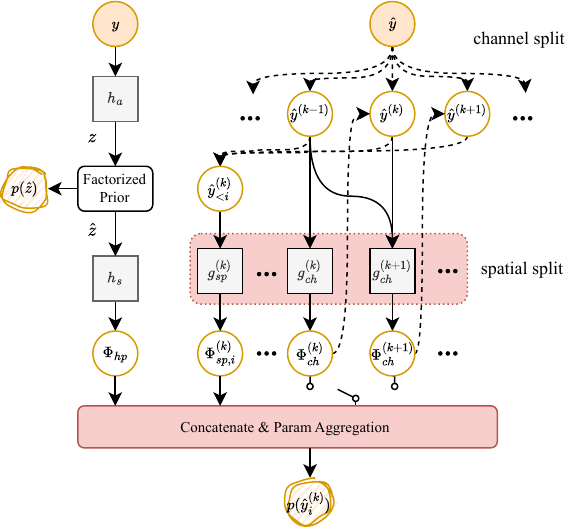}
    \caption{Hyperprior and Spatial-Channel Context (SCCTX) entropy model for patch-token coding. The hyperprior provides side information, while spatial and channel contexts are combined to predict local Gaussian parameters for entropy coding~\cite{He.2022.ELIC}.}
    \label{fig:scctx}
\end{figure}

The spatial, channel, and hyperprior contexts are concatenated channel-wise and fed into a parameter network $g_{\mathrm{ep}}$ to predict location-wise Gaussian parameters:
\begin{equation}
    \boldsymbol{\Theta}_i^{(k)} = g_{\mathrm{ep}}\!\left(\boldsymbol{\Phi}_{\mathrm{sp},i}^{(k)},\, \boldsymbol{\Phi}_{\mathrm{ch},i}^{(k)},\, \boldsymbol{\Phi}_{\mathrm{hp},i}\right) = \bigl(\boldsymbol{\mu}_i^{(k)}, \boldsymbol{\sigma}_i^{(k)}\bigr),
\end{equation}
where $i$ indexes spatial positions.
Using these parameters, the latent $\mathbf{y}_i^{(k)}$ is quantized as:
\begin{equation}
    \hat{\mathbf{y}}_i^{(k)} = \operatorname{round}\!\bigl(\mathbf{y}_i^{(k)} - \boldsymbol{\mu}_i^{(k)}\bigr) + \boldsymbol{\mu}_i^{(k)}.
\end{equation}
Once decoded, $\hat{\mathbf{y}}_i^{(k)}$ updates the context for subsequent coding steps, iterating until all latents are encoded.

By concatenating all local entropy parameters, we model the overall distribution of $\hat{\mathbf{y}}$ given the hyperprior $\hat{\mathbf{z}}$ as a conditional Gaussian:
\begin{equation}
    p_{\hat{\mathbf{y}} \mid \hat{\mathbf{z}}}(\hat{\mathbf{y}} \mid \hat{\mathbf{z}}) \sim \mathcal{N}\!\bigl(\boldsymbol{\mu}, \operatorname{diag}(\boldsymbol{\sigma}^2)\bigr),
\end{equation}
and the resulting rate for $\hat{\mathbf{y}}$ is given by:
\begin{equation}
    R(\hat{\mathbf{y}}) = \mathbb{E}\left[-\log_2 p_{\hat{\mathbf{y}} \mid \hat{\mathbf{z}}}(\hat{\mathbf{y}} \mid \hat{\mathbf{z}})\right].
\end{equation}

\subsection{Rate–Distortion Optimization}
\label{subsec:rd}

Previous feature compression methods typically measure distortion at the compressed layer:
\begin{equation}
    \mathcal{L}_{\text{D}}^{\text{base}} = \|\mathbf{t}_n - \hat{\mathbf{t}}_n\|_2^2,
\end{equation}
which penalizes token-wise error but does not constrain how quantization noise propagates through later ViT blocks.
Small compressed-layer MSE can still yield large drift in deep features.

\textbf{Feature-matching loss.} We supervise the compression using deeper features of the subsequent $j$ blocks after the compressed layer
\begin{equation}
    \begin{aligned}
        \mathbf{t}_{n+j}       & = \mathrm{Block}_{n+1:n+j}(\mathbf{t}_n),       \\
        \hat{\mathbf{t}}_{n+j} & = \mathrm{Block}_{n+1:n+j}(\hat{\mathbf{t}}_n),
    \end{aligned}
\end{equation}

When calculating loss, we decompose them into patch tokens $\mathbf{t}_{n+j}^{(2d)}$ and global tokens $\mathbf{t}_{n+j}^{(1d)}$, and define the \textbf{feature-matching} distortion term as:
\begin{equation}
    \mathcal{L}_{\text{D}} = \|\mathbf{t}_{n+j}^{(2d)} - \hat{\mathbf{t}}_{n+j}^{(2d)}\|^2_2 + \|\mathbf{t}_{n+j}^{(1d)} - \hat{\mathbf{t}}_{n+j}^{(1d)}\|^2_2,
\end{equation}

Finally, we optimize a rate–distortion objective that accounts for the bit‑rates of global-token latents $\hat{\mathbf{e}}$, hyper latents $\hat{\mathbf{z}}$, and patch latents $\hat{\mathbf{y}}$:
\begin{equation}
    \mathcal{L} = \mathcal{L}_{\text{D}} + \lambda \cdot \bigl(R(\hat{\mathbf{e}})+R(\hat{\boldsymbol{z}})+R(\hat{\boldsymbol{y}})\bigr)
    = \mathcal{L}_{\text{D}} + \lambda \cdot R_{\mathrm{total}},
    \label{eq:rd_loss}
\end{equation}
where $\lambda$ is the Lagrangian multiplier to control the rate-distortion trade-off.
The same $\lambda$ weights both the 1D (global) and 2D (patch) rate terms.

\textbf{Balancing global and patch tokens.}
A natural concern is that global tokens are far fewer than patch tokens and might be under-optimized when using the same $\lambda$ for both branches.

Equivalently, with $R_{1\mathrm{d}} = R(\hat{\mathbf{e}})$ and $R_{2\mathrm{d}} = R(\hat{\boldsymbol{z}})+R(\hat{\boldsymbol{y}})$,
\begin{equation}
    \mathcal{L} = (\lambda \cdot R_{1\mathrm{d}} + D_{1\mathrm{d}}) + (\lambda \cdot R_{2\mathrm{d}} + D_{2\mathrm{d}}),
\end{equation}
where $D_{1\mathrm{d}}$ and $D_{2\mathrm{d}}$ denote the global- and patch-token parts of $\mathcal{L}_{\text{D}}$.

Let $N$ denote the total number of supervised tokens, with $a$ global tokens and $N-a$ patch tokens ($N \gg a$).
A token-count--weighted MSE would be
$\mathrm{MSE}_{\mathrm{total}} = \frac{a}{N}\mathrm{MSE}_{1\mathrm{d}} + \frac{N-a}{N}\mathrm{MSE}_{2\mathrm{d}}$,
whereas our distortion uses $\mathcal{L}_{\text{D}} = \mathrm{MSE}_{1\mathrm{d}} + \mathrm{MSE}_{2\mathrm{d}}$.
This assigns a much higher effective weight to global-token error relative to their count, encouraging accurate reconstruction of semantically critical global tokens.
The $1{:}1$ weighting between $\mathrm{MSE}_{1\mathrm{d}}$ and $\mathrm{MSE}_{2\mathrm{d}}$ is an empirical choice that jointly supports classification (global-heavy) and segmentation (patch-heavy).

\subsection{Variable-Rate Model}

To enable flexible deployment, we integrate a \textit{single-model variable-rate} mechanism~\cite{Jia.2025.DCVC-RT,Xu.2025.MNI,Zhang.2025.LSP}. Learning-based codecs often train separate models per rate point; instead, we condition the same weights on a quantization parameter QP that selects a Lagrange multiplier $\lambda$ in Eq.~\eqref{eq:rd_loss}. The core idea is that, for a fixed quantization step size, scaling the magnitude of the latent $\mathbf{y}$ changes the output bitrate. To allow a single model to support multiple rates, we condition it on the Lagrange multiplier $\lambda$, which governs the rate-distortion trade-off during training.

Concretely, we predefine a set of $\lambda$ coefficients, each corresponding to a quantization parameter (QP) value. During forward propagation, the input QP is mapped through an embedding layer to four modulation vectors $\mathbf{q}_1$ to $\mathbf{q}_4$, as depicted in the module bank of Fig.~\ref{fig:codec}. These vectors perform channel-wise feature weighting, thereby adjusting the effective information content and enabling variable-rate coding within a single model.

\section{Experiments}
\label{sec:exp}


\subsection{Experimental Setup}
\label{subsec:experimental_setup}

We validate our Visual Token Codec (VTC) on two ViT backbone families---DINOv2 and SAM3---with frozen backbones and downstream heads.
Each codec is trained once per backbone size and evaluated with frozen backbone weights and frozen task heads.

\subsubsection{SAM3 Backbone}
\label{subsubsec:sam3_setup}

We use publicly released pretrained SAM3 weights~\cite{Carion2025SAM3} and compress intermediate tokens from the \emph{image encoder} after its 32 ViT blocks, before three parallel multi-scale convolutional heads.
Reconstructed tokens pass through these heads and the remaining detector/segmentation modules.
For SAM3, VTC encodes only patch tokens; the latent dimension is reduced to 48, with SCCTX using three channel- and two spatial-autoregressive steps.

\subsubsection{DINOv2 Backbones}
\label{subsubsec:dinov2_setup}

We evaluate our method on both the standard DINOv2 and its register-enhanced variant (DINOv2 Reg4)~\cite{Darcet.2024.ViTRegisters,Oquab2023DINOv2LR}, implemented via the timm library with patch size 16. We compress last-layer tokens after all ViT blocks and before the last norm layer. For multi-task-compatible settings, we compress from the third-to-last layer (Layer~09), reconstruct the last four layers after decoding, and jointly feed them to downstream task heads for better performance.

For the small (S), base (B), and large (L) variants of the DINOv2 backbone, the feature dimensions are 384, 768, and 1024, respectively. In the codec, we reduce the features to latent dimensions of 48, 48, and 120, respectively.
Note that the Large model keeps a higher dimension to avoid an overly strict information bottleneck.
The SCCTX entropy model performs 3, 3, and 5 channel-wise autoregressive steps for S, B, and L models, respectively.

\subsubsection{Training Protocol}
\label{subsubsec:training}

DINOv2 codecs are trained on the ImageNet-1K training split~\cite{Russakovsky2015ImageNet} for 20 epochs using Adam on a single NVIDIA RTX~4090. We supervise MSE on the final outputs of the last normalization layer.
We apply random resized-crop augmentation during training, and feature targets are extracted online from frozen DINOv2 weights.

SAM3 codecs are trained on the COCO2017 training split~\cite{Lin.2014.MSCOCO}, following the same procedure. They supervise MSE on the final multi-scale convolutional outputs, with a scalar $\alpha$ balancing scale-wise loss magnitudes.

For variable-rate coding, we first pretrain a single-rate model at high bitrate and then continue VBR training from that checkpoint. During VBR training, each iteration uniformly samples a QP and its paired Lagrange multiplier $\lambda$. Specifically, $\lambda$ is sampled from 65 logarithmically spaced values in $[1, 256]$, corresponding to QP values in $[0, 64]$.

\subsubsection{Baselines}
\label{subsubsec:baselines}

We compare VTC with an uncompressed upper bound and two feature-coding baselines.
\textbf{Bypass} directly uses uncompressed intermediate features and serves as the performance upper bound.
\textbf{LaMoFC-VTM}~\cite{Gao.2025.LaMoFC} is the primary baseline: it reshapes the token sequence into a pseudo image, clips and quantizes the values to a 10-bit range, and compresses the result with VTM~\cite{Bross2021VTM}.
We do not report LaMoFC-Hyperprior because it performs worse than the VTM-based variant in the original study.
\textbf{VQFC}~\cite{Chen.2026.VQFC} is additionally included: it partitions the feature tensor into 2D chunks and maps each chunk to a shared codebook.

\subsubsection{Evaluation Protocols}
\label{subsubsec:eval_protocols}

Table~\ref{tab:eval_setup} summarizes the per-task settings used in our evaluation, including the backbone, task, preprocessing, feature shape, and test dataset.

\paragraph{Task-specific setting.}
This setting covers SAM3 Det/InsSeg and DINOv2 Cls/SemSeg/Rec.
For SAM3, images are resized to $1008{\times}1008$, producing a fixed $72{\times}72$ patch-token grid, and features are extracted from the specified ViT layer of the image encoder.
For DINOv2-B/16-Reg4, inputs follow the task-specific preprocessing in Table~\ref{tab:eval_setup}: Cls uses resized $256{\times}256$ images, SemSeg uses resized inputs with sliding-window inference, and Rec uses padded inputs.
Because DINOv2-Reg4 contains one CLS token and four register tokens in addition to patch tokens, the feature shapes include five extra tokens, e.g., $261{=}16{\times}16{+}5$ for $256{\times}256$ classification inputs and $(H{\times}W/256{+}5)$ for padded variable-size inputs.

\paragraph{Multi-task compatible setting.}
This setting uses only simple padding (\texttt{PadMultiple}) without task-specific resizing or sliding-window extraction.
The resulting features preserve the input image layout up to padding and can therefore be reused by different downstream tasks from a single compressed bitstream.

\paragraph{Rate metrics.}
For task-specific fixed-resolution protocols, we follow LaMoFC~\cite{Gao.2025.LaMoFC} and report \textit{bits per feature point} (bpfp), which normalizes bitrate by the number of feature points.
For the multi-task compatible protocol, we report \textit{bits per pixel} (bpp), where bitrate is normalized by the original image resolution and therefore better reflects practical bandwidth and storage costs.

\paragraph{Task metrics and BD analysis.}
Cls reports top-1 accuracy, SemSeg reports mIoU on ADE20K~\cite{Zhou2017ADE20K}, and SAM3 Det/InsSeg report COCO-style box mAP and mask AP on MSCOCO~\cite{Lin.2014.MSCOCO}, respectively.
To quantify coding efficiency, we use Bjontegaard-Delta (BD) metrics~\cite{bd-metrics}, which require sufficient overlap between two rate--performance curves.
\textbf{BD-Rate} (lower is better) measures the average bitrate reduction at equivalent performance, while \textbf{BD-Acc}, \textbf{BD-mIoU}, and \textbf{BD-mAP} (higher is better) measure the average performance gain at equivalent bitrate.
We report BD metrics only when curves overlap sufficiently; otherwise, entries are marked ``--''.

\paragraph{Test Dataset}
Prior work argues that small subsets (e.g., 100 or 500 images) suffice to reveal compression trends.
We find such tiny subsets unstable, as they often yield noisy rate--performance curves for complex models.
Therefore, we use more than 1000 images for each task whenever possible (Table~\ref{tab:eval_setup}).
Most tasks directly use the corresponding validation split; for ImageNet, we use a 2K-image subset selected from the validation split, denoted as ImageNet sel2k.

\begin{table*}[t]
\centering
\caption{Per-task experimental settings for SAM3-L and DINOv2-B-Reg4, including input preprocessing, feature tensor shape, and evaluation data.}
\label{tab:eval_setup}

\begin{tabular}{lllll}
    \hline
                                                                                            & Task                            & Preprocess         & Feature Shape      & Test Dataset          \\ \hline
    \multirow{2}{*}{\begin{tabular}[c]{@{}l@{}}SAM3-L/14\\ (32 layers)\end{tabular}}        & Object Detection (Det)          & Resize(1008,1008)  & $5184, 1024$       & MSCOCO 2017 val, 5k   \\
                                                                                            & Instance Segmentation (InsSeg)  & Resize(1008,1008)  & $5184, 1024$       & MSCOCO 2017 val, 5k   \\ \hline
    \multirow{4}{*}{\begin{tabular}[c]{@{}l@{}}DINOv2-B/16-reg4\\ (12 layers)\end{tabular}} & Image Classification (Cls)      & Resize(256,256)    & $261, 768$         & ImageNet selected, 2k \\
                                                                                            & Semantic Segmentation (SemSeg)  & Resize(512), Slide & $n, 1029, 768$     & ADE20K val, 2k        \\
                                                                                            & Subjective Reconstruction (Rec) & PadMultiple        & $(H*W/256+5), 768$ & Any                   \\
                                                                                            & Multi-Task                      & PadMultiple        & $(H*W/256+5), 768$ & Any                   \\ \hline
    \end{tabular}

\end{table*}

\subsection{Main Results}
\label{subsec:main_results}

We first quantify coding performance under increasingly general evaluation protocols.
Unless otherwise stated, all comparisons use the baselines defined in Sec.~\ref{subsubsec:baselines}.

\subsubsection{Task-Specific Feature Coding}
\label{subsubsec:task_specific}

Most existing ViT feature coding methods focus on compressing final-layer features under task-specific preprocessing (Table~\ref{tab:eval_setup}). 

\begin{figure*}[ht]
    \centering
    \includegraphics[width=0.48\textwidth]{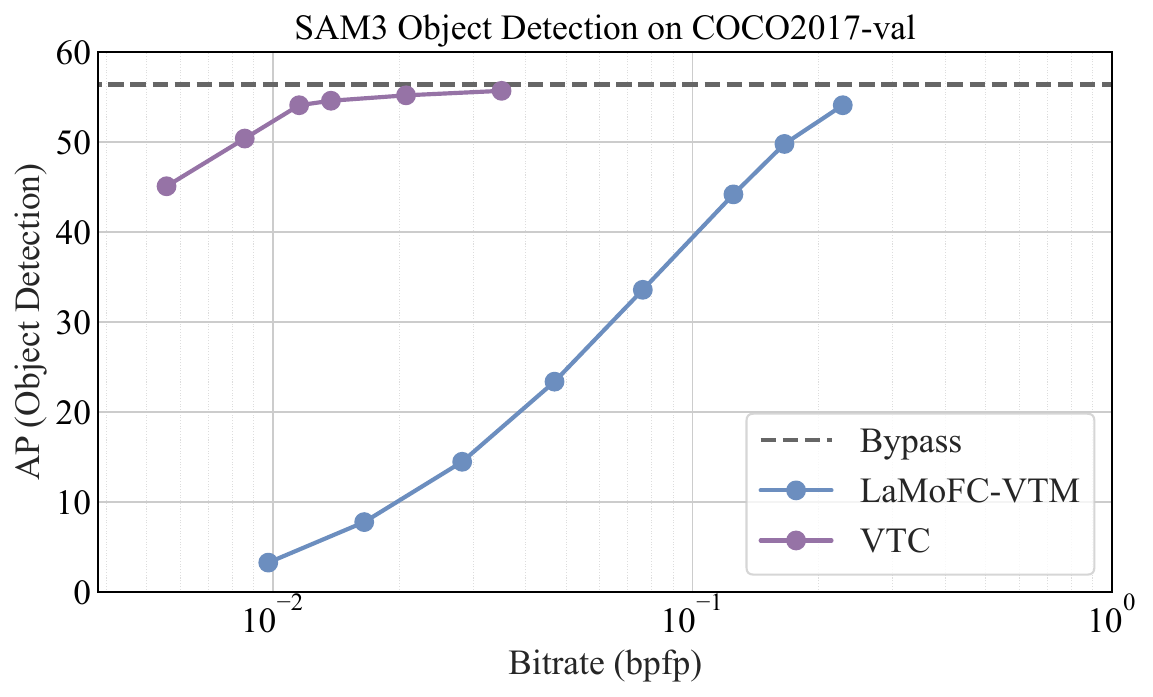}\hfill
    \includegraphics[width=0.48\textwidth]{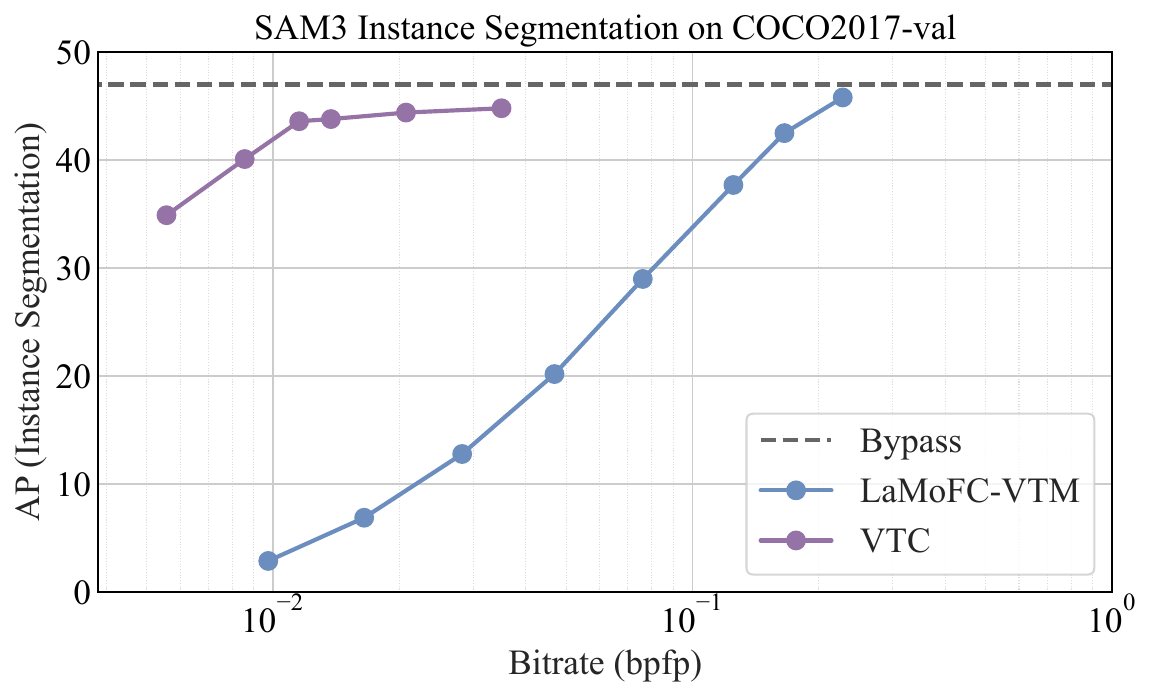}
    \\[0.5em]
    \includegraphics[width=0.48\textwidth]{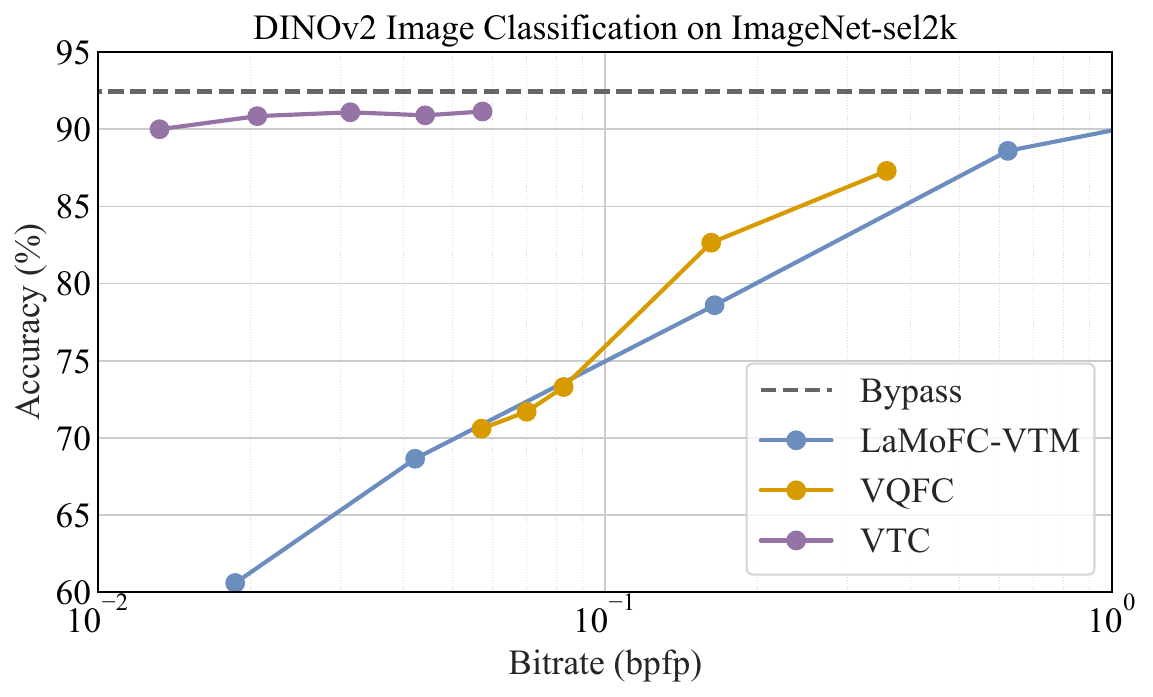}\hfill
    \includegraphics[width=0.48\textwidth]{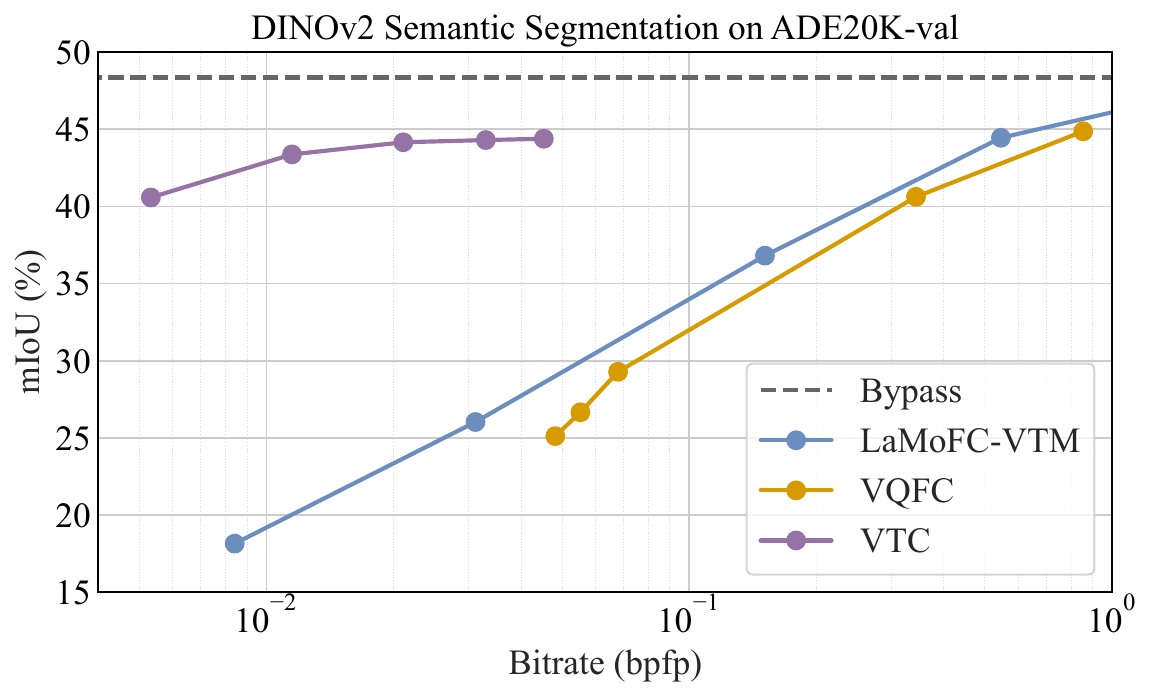}
    \caption{Rate--performance under task-specific feature coding for two foundation models and four tasks. Bypass: uncompressed features; LaMoFC VTM: VTM-compressed features; VTC: our compressed features. \textbf{Top-left:} SAM3 object detection (Det). \textbf{Top-right:} SAM3 instance segmentation (InsSeg). \textbf{Bottom-left:} DINOv2 image classification (Cls). \textbf{Bottom-right:} DINOv2 semantic segmentation (SemSeg).}
    \label{fig:fixed_tasks}
\end{figure*}

Fig.~\ref{fig:fixed_tasks} reports rate--performance curves on SAM3 Det/InsSeg and DINOv2 Cls/SemSeg.
Because the curves of different methods have limited overlap, BD metrics cannot be computed reliably here; we therefore compare the bitrate required to reach 90\% of the bypass (uncompressed-feature) performance (Table~\ref{tab:task_specific_90}).

\begin{table}[t]
    \centering
    \caption{Bitrate comparison at 90\% of bypass performance under task-specific protocols. Bitrate is reported in bpfp.}
    \label{tab:task_specific_90}
    \begin{tabular}{@{}lrrrr@{}}
    \toprule
    Task & Performance & LaMoFC VTM & VTC & Relative \\ \midrule
    SAM3 Det    & AP@50.8   & 0.179 & 0.009 & 20.3$\times$ \\
    SAM3 InsSeg & AP@42.3   & 0.164 & 0.010 & 15.7$\times$ \\ \midrule
    DINOv2 Cls  & Acc@83.2  & 0.375 & $\leq$0.013 & 28.4$\times$ \\
    DINOv2 SemSeg & mIoU@43.5 & 0.498 & 0.013 & 37.4$\times$ \\ \bottomrule
    \end{tabular}
    \end{table}

Across all four tasks, VTC substantially reduces the bitrate needed to reach 90\% of bypass performance, with relative gains ranging from 15.7$\times$ (SAM3 InsSeg) to 37.4$\times$ (DINOv2 SemSeg).
On SAM3, VTC reaches the Det target at 0.009~bpfp versus 0.179~bpfp for LaMoFC-VTM, suggesting that directly applying image codecs to reshaped ViT tokens is inefficient for detection-oriented features.
InsSeg remains more sensitive than Det because mask prediction depends more strongly on spatial and multi-scale details, yet VTC still preserves task-relevant token structure at much lower bitrate.
On DINOv2, VTC remains close to bypass accuracy well into the low-bitrate region for Cls, whereas VTM-based coding degrades much earlier.
The largest gain on SemSeg indicates that explicit 2D patch-token modeling is especially important for dense prediction tasks.

\subsubsection{Multi-Task Compatible Coding}
\label{subsubsec:multitask}

\begin{figure*}[t]
    \centering
    \includegraphics[width=0.48\linewidth]{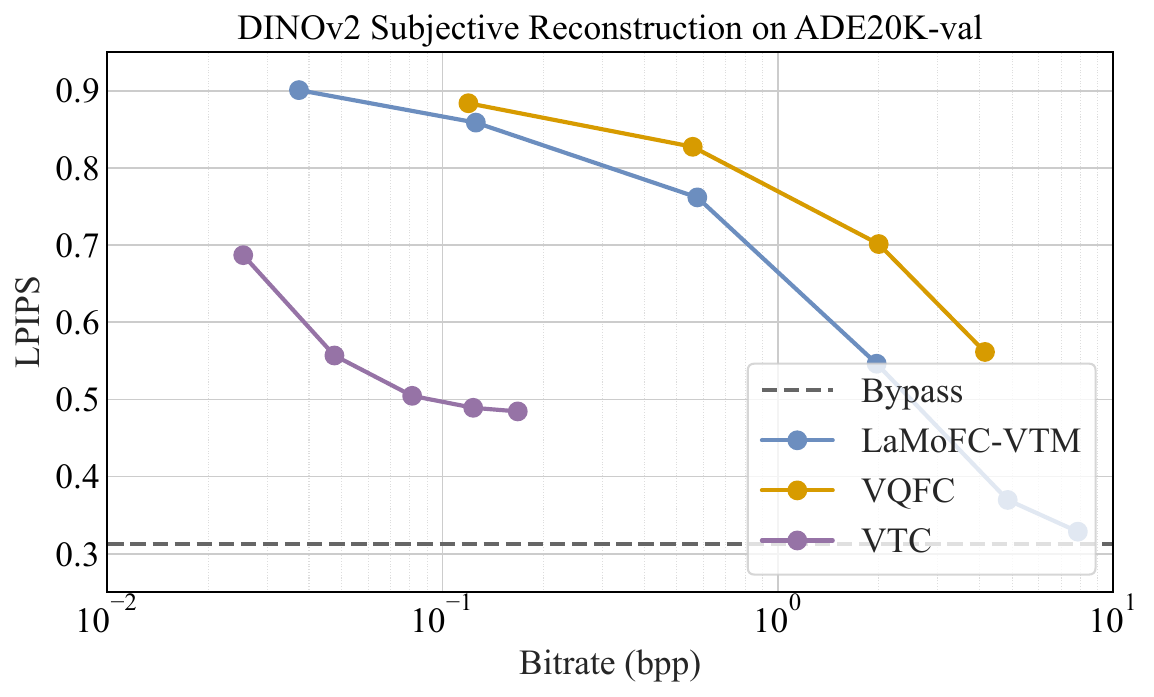}
    \hfill
    \includegraphics[width=0.48\textwidth]{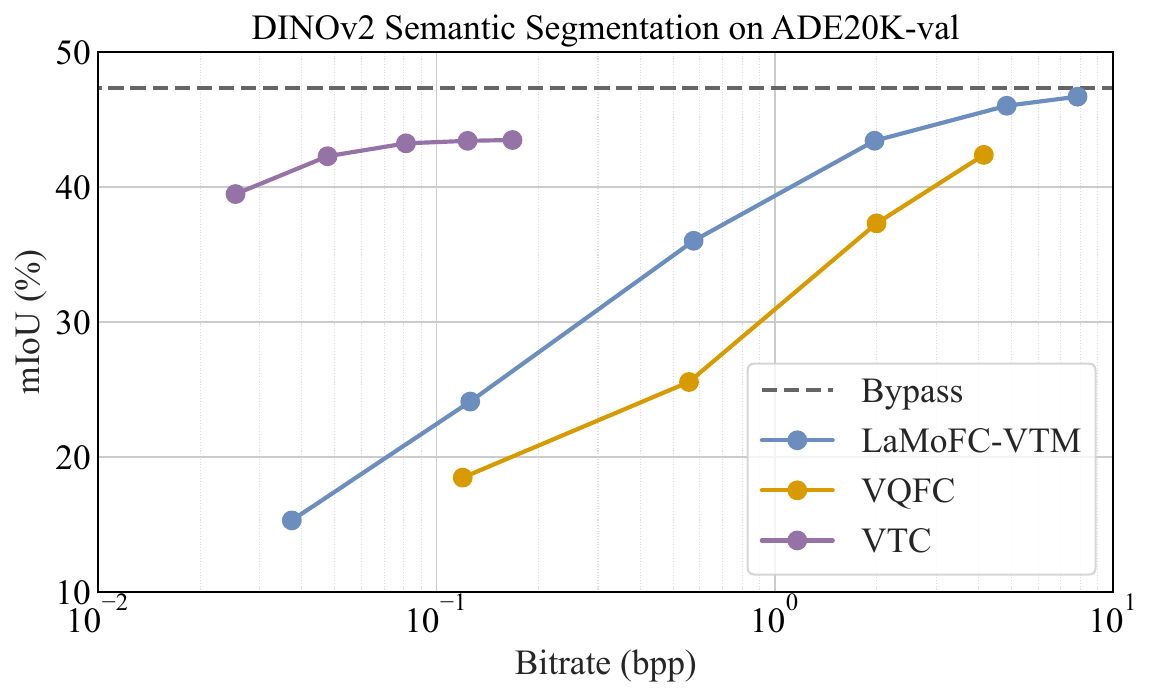}
    \hfill
    \caption{Rate--performance under the multi-task-compatible setting using DINOv2-Base-Reg4 features on ADE20K val. Bypass: uncompressed features; LaMoFC VTM: VTM-compressed features; VTC: our compressed features. \textbf{Left:} subjective reconstruction measured by LPIPS. \textbf{Right:} semantic segmentation measured by mIoU.}
    \label{fig:varia_tasks}
\end{figure*}

We next evaluate whether a \emph{single} compressed feature bitstream can support multiple downstream tasks without task-specific resizing.
Under the multi-task-compatible protocol, images keep their original resolution with only minimal padding (\texttt{PadMultiple}), and bitrate is reported in bpp.
The same DINOv2-Base-Reg4 bitstream is used for subjective reconstruction (Rec) and semantic segmentation (SemSeg) on ADE20K val~\cite{Zhou2017ADE20K} (Fig.~\ref{fig:varia_tasks}).

For SemSeg, VTC maintains a clear advantage over LaMoFC-VTM, consistent with the task-specific results.
We also reproduce VQFC under this variable-resolution setting but observe worse performance than LaMoFC-VTM, likely because VQFC was originally designed for fixed-resolution feature tensors and does not adapt well to variable-size token maps.
For Rec, decoded features are fed to a frozen RAE decoder~\cite{RAE} without fine-tuning and evaluated with LPIPS.
Although VTC does not reach a near-lossless reconstruction regime within the evaluated rate points, it uses 17.7$\times$ less bitrate than LaMoFC-VTM at the same LPIPS$=$0.484, while still leaving a gap to bypass reconstruction.
This highlights a practical trade-off: task-oriented coded features preserve semantics needed for Seg, but may lose pixel-level details less critical for dense prediction.

\subsection{Intermediate-Layer Analysis for Deployment}
\label{subsec:intermediate_layer}

The compressed layer determines which intermediate features are coded, thereby coupling rate--distortion behavior with the sender/receiver computation budget in distributed deployment (Fig.~\ref{fig:framework}).
We analyze compressed-layer choices on DINOv2-B-Reg4 with ADE20K SemSeg under the multi-task-compatible protocol.
\textbf{Input} denotes the ViT layer whose features are compressed, and \textbf{Spv} denotes the supervision layer used in the feature-matching loss.
After intermediate features are decompressed, the decoder reconstructs the remaining layers and feeds them jointly to the task head.

\begin{figure*}[t]
    \centering
    \includegraphics[width=0.48\linewidth]{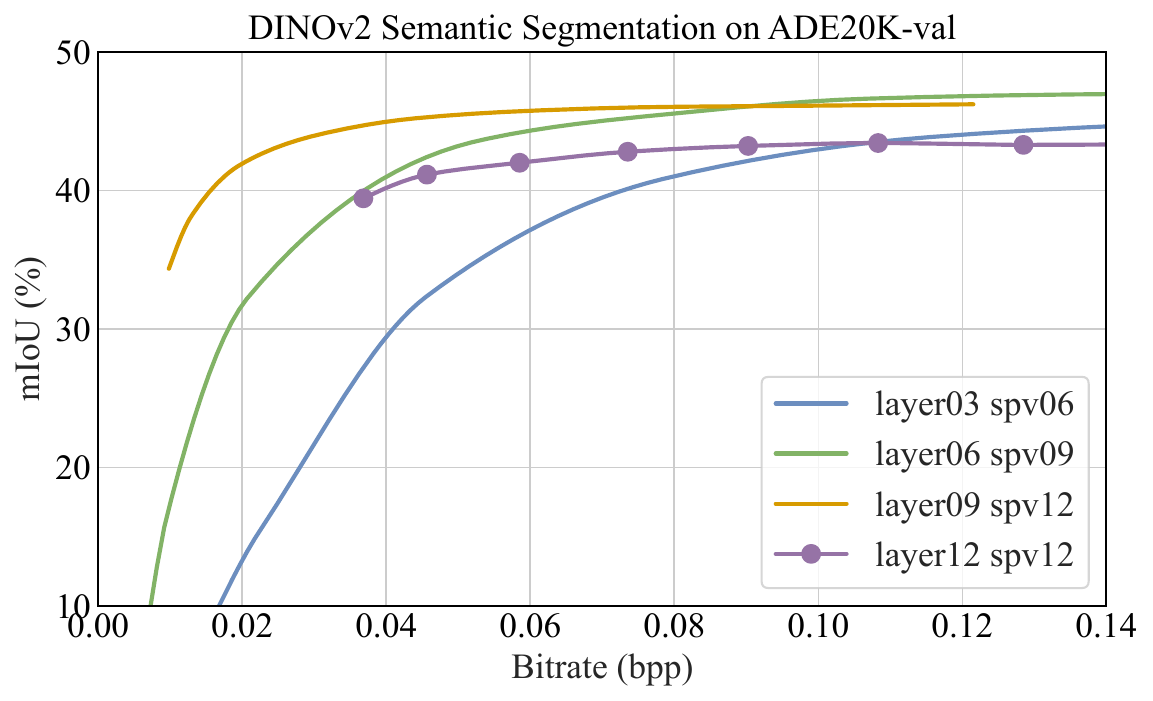}
    \hfill
    \includegraphics[width=0.48\textwidth]{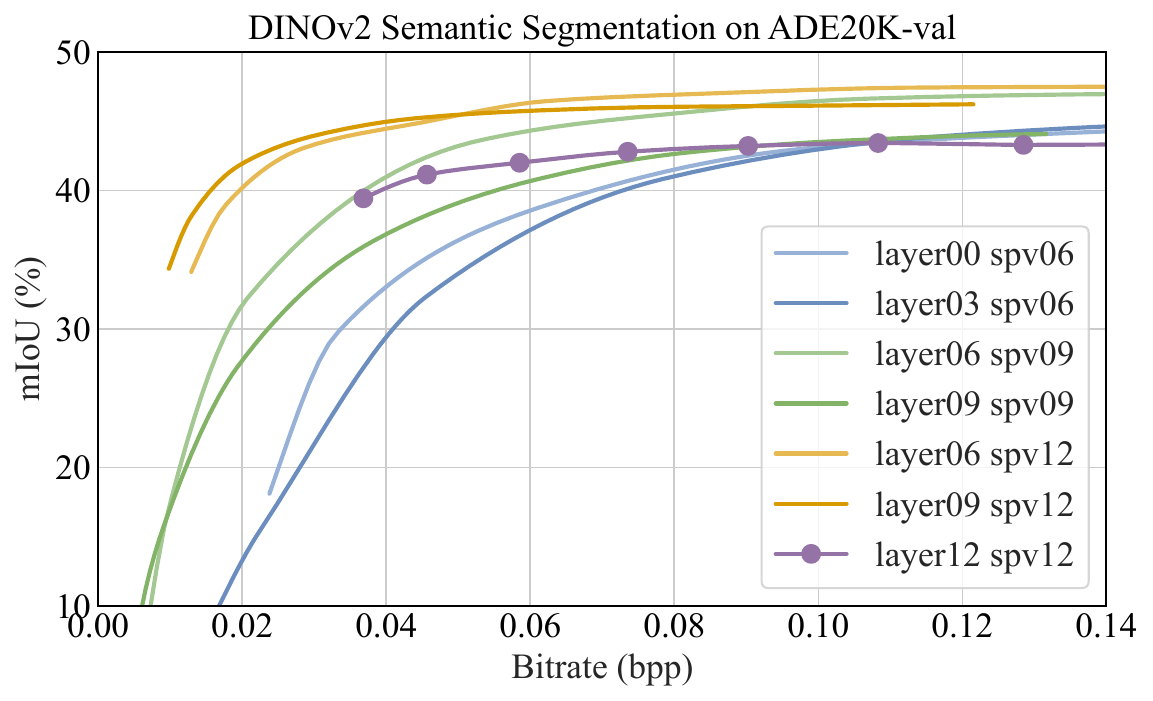}
    \hfill
    \caption{Rate--performance curves under different compressed-layer choices on ADE20K semantic segmentation. Left: varying the compressed input layer. Right: varying the feature-matching supervision layer.}
    \label{fig:split}
\end{figure*}

Fig.~\ref{fig:split} first examines the rate--performance effect of changing \textbf{Input} and \textbf{Spv}.
Among the tested compressed layers, deeper input features are generally easier to compress than early features, since they contain richer semantics and less low-level detail.
However, directly compressing the final layer is not optimal for ADE20K SemSeg: compressing Layer~9 and reconstructing the last four layers provides stronger dense-prediction cues than coding the final representation alone.
The supervision layer further shapes the RD trend, as curves with different \textbf{Input} layers but the same \textbf{Spv} layer exhibit similar trajectories.
This suggests that feature-matching supervision largely determines what information the codec is encouraged to preserve.

Table~\ref{tab:ablation} further relates these RD trends to system cost.
BD metrics are computed against the default multi-task operating point in row~3 (Input$=$9, Spv$=$12, $f_a{=}f_s{=}2$).
Rows~1--3 vary the compressed input layer with the same codec-transform depth, rows~6--7 represent extreme compressed-layer choices, and rows~4--5 are included for the codec-block ablation later discussed in Sec.~\ref{subsec:ablation}.
FLOPs are measured on $512{\times}512$ images and divided into backbone \textbf{prefix}, entropy \textbf{compress}/\textbf{decompress}, and backbone+codec \textbf{suffix}.

\begin{table*}[t]
\centering
\caption{Compressed-layer trade-offs on ADE20K SemSeg under the multi-task-compatible protocol. We report FLOPs measured on 512$\times$512 images and BD metrics relative to row~3. Encoding comprises \textbf{prefix} and \textbf{compress} stages; decoding comprises \textbf{decompress} and \textbf{suffix} stages. Rows~4--5 are later discussed as codec-transform ablations in Sec.~\ref{subsec:ablation}.}
\label{tab:ablation}
\begin{tabularx}{\textwidth}{@{} 
    *{4}{>{\raggedleft\arraybackslash\hsize=0.5\hsize}X}
    *{6}{>{\raggedleft\arraybackslash\hsize=1.3\hsize}X}
    @{}}
\toprule
\multicolumn{2}{c}{Layer index} & \multicolumn{2}{c}{\#Blocks} & \multicolumn{2}{c}{Sender FLOPs (G)} & \multicolumn{2}{c}{Receiver FLOPs (G)} & \multicolumn{2}{c}{Performance} \\
\cmidrule(r){1-2} \cmidrule(lr){3-4} \cmidrule(lr){5-6} \cmidrule(lr){7-8} \cmidrule(l){9-10}
Input & Spv & $f_a$ & $f_s$ & prefix & compress & decompress & suffix & BD-rate & BD-mIoU \\
\midrule
3 & 6 & 2 & 2 & 22.477 & 15.099 & 15.089 & 65.635 & 357.88\% & -17.99 \\
6 & 9 & 2 & 2 & 44.350 & 15.099 & 15.089 & 43.762 & 111.22\% & -6.33 \\

\rowcolor{gray!15}
9 & 12 & 2 & 2 & 66.223 & 15.099 & 15.089 & 21.889 & 0.00\% & 0.00 \\

\midrule
9 & 12 & 1 & 1 & 66.223 & 7.808 & 7.798 & 21.889 & 55.30\% & -2.49 \\
9 & 12 & 0 & 0 & 66.223 & 0.517 & 0.507 & 21.889 & -- & -15.51 \\
\midrule
12 & 12 & 2 & 2 & 88.097 & 15.099 & 15.089 & 0.004 & 180.08\% & -3.54 \\
0 & 12 & 0 & 2 & 0.604 & 0.517 & 15.089 & 87.508 & 404.89\% & -5.26 \\
\bottomrule
\end{tabularx}

\end{table*}

\begin{figure*}[ht!]
    \centering
    \includegraphics[width=0.5\textwidth]{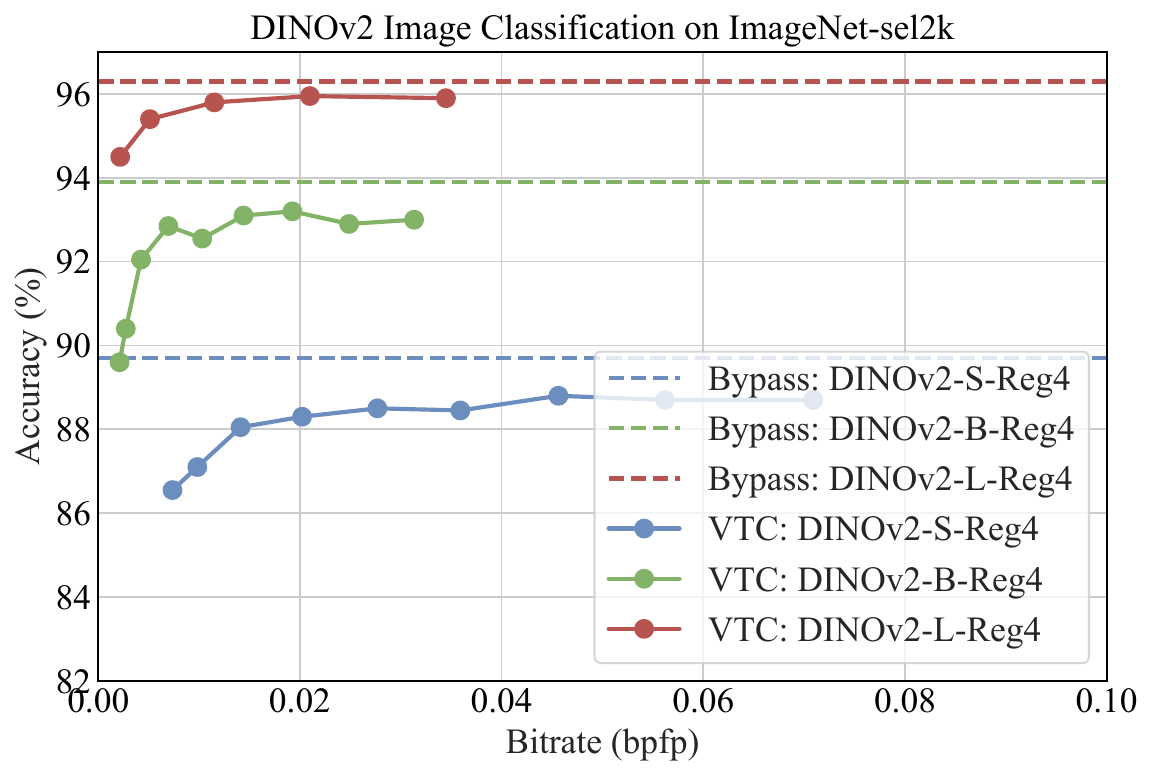}\hfill
    \includegraphics[width=0.5\textwidth]{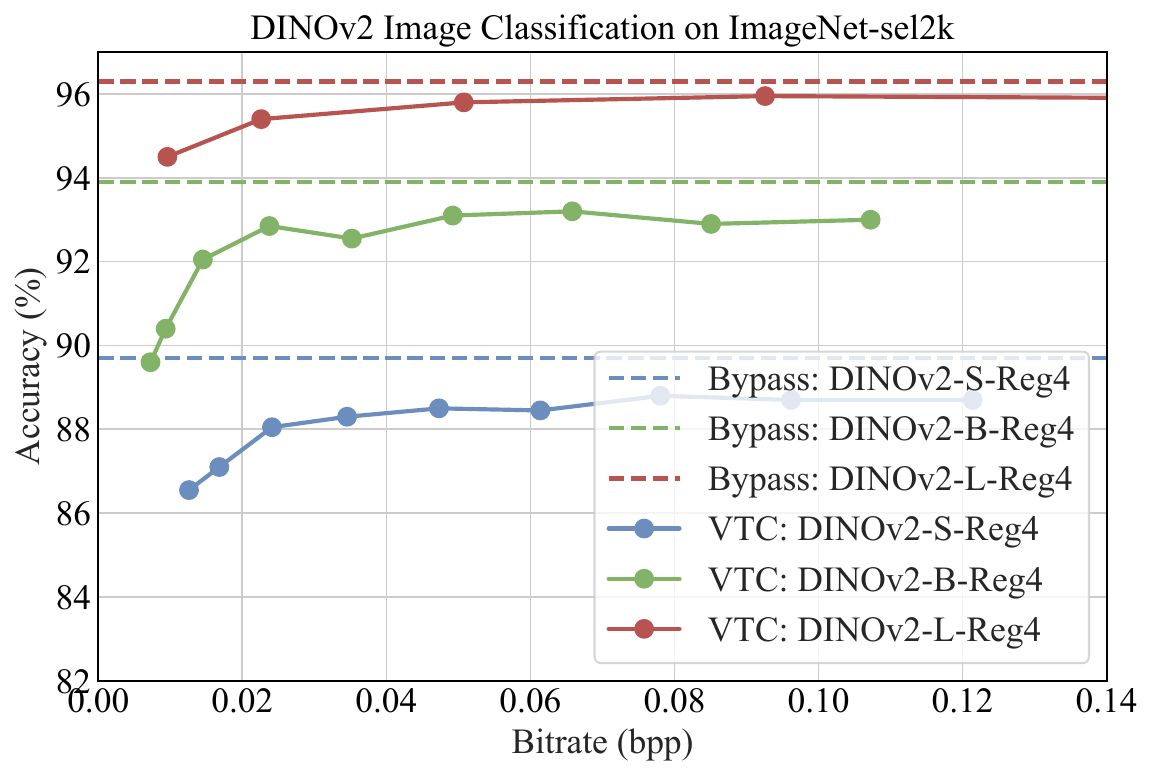}\hfill
    \\[0.5em]
    \includegraphics[width=0.5\textwidth]{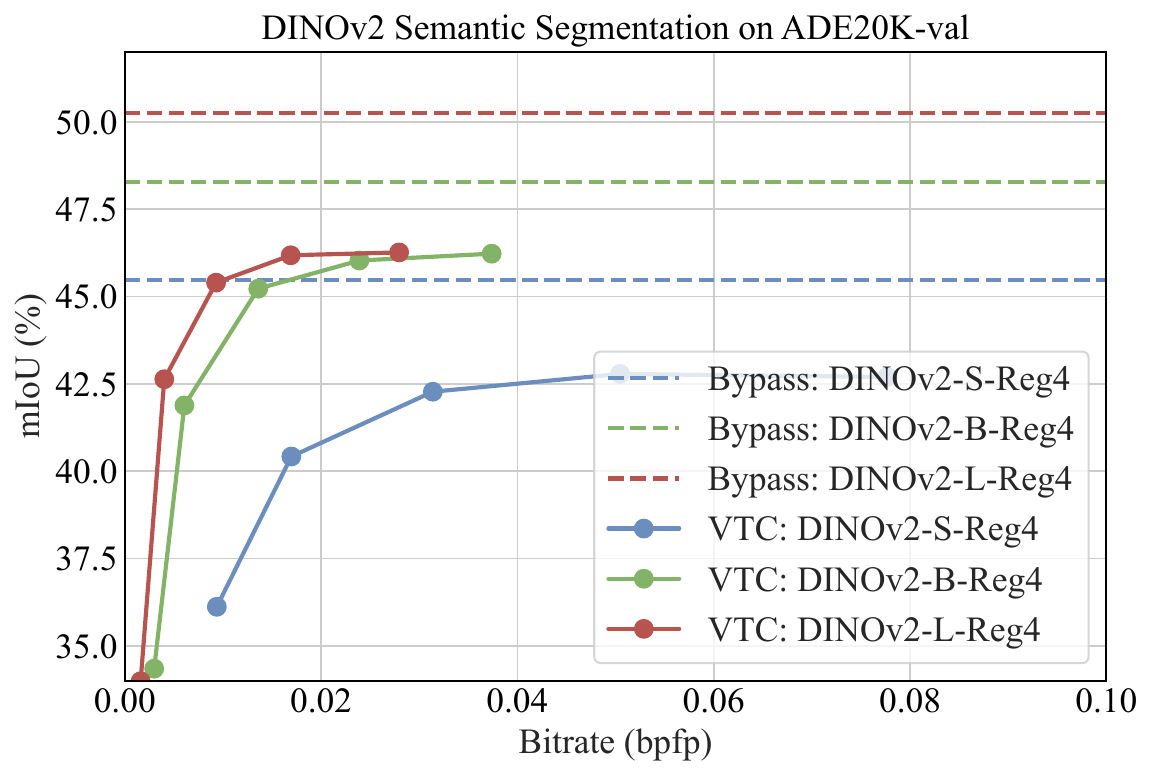}\hfill
    \includegraphics[width=0.5\textwidth]{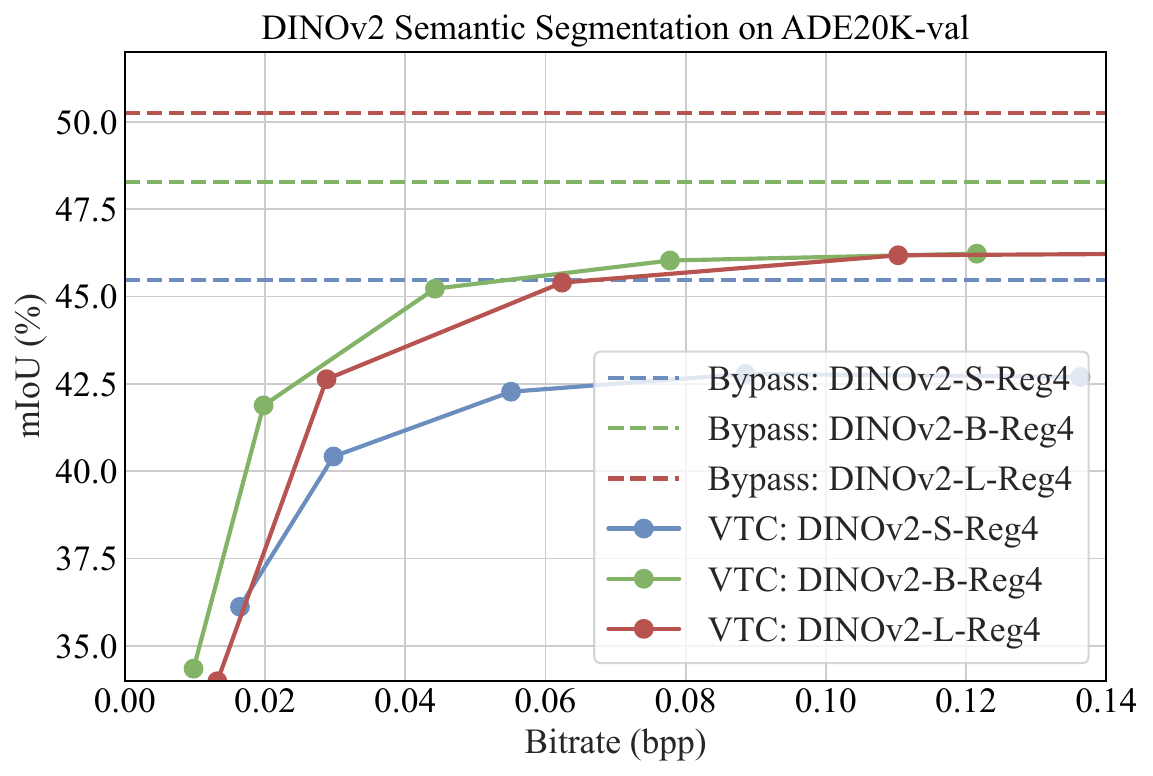}\hfill
    \caption{Rate--performance curves for VTC with different DINOv2-Reg4 backbone scales. Horizontal lines indicate uncompressed baselines. Top: ImageNet sel2k classification; bottom: ADE20K val segmentation. Left: bpfp; right: bpp.}
    \label{fig:scale}
\end{figure*}

These operating points clarify how different compressed layers instantiate the two deployment paradigms in Fig.~\ref{fig:framework}.

\textbf{Transmission-oriented feature coding.}
For one-time, task-specific edge--cloud inference, compressing shallow features is attractive because it limits sender-side computation while offloading most ViT processing to the receiver.
Row~7 gives an extreme shallow-layer example: the prefix costs only $\sim$0.6\,G FLOPs, but the suffix grows to $\sim$87.5\,G FLOPs and the RD penalty becomes large.
This operating point is therefore suitable when reducing on-device computation is more important than achieving the best compression efficiency.

\textbf{Storage-oriented feature coding.}
For large-scale feature banks, retrieval, or repeated offline analysis, compressing deeper features is more natural because the expensive prefix can be executed once and the compressed representation can be stored and reused by different downstream tasks.
Row~6 (Input$=$12) illustrates this regime: the prefix costs $\sim$88\,G FLOPs, but the suffix is almost negligible ($\sim$0.004\,G FLOPs). Its 180.08\% BD-Rate trade-off suggests that final-layer storage is not always the most RD-efficient point, but it aligns with the storage-oriented goal of producing high-level reusable representations.
Overall, row~3 provides the best RD efficiency among the tested configurations, while shallow and deep compressed layers expose explicit computation--bandwidth operating points for different deployment constraints.

\subsection{Backbone Scale}
\label{subsec:more_dinov2_results}

DINOv2 provides multiple backbone scales distilled from larger teacher models. To study how backbone capacity affects feature coding, we evaluate VTC on the DINOv2-S/B/L-Reg4 family under the same protocol.

As shown in Fig.~\ref{fig:scale}, increasing the backbone scale generally improves the achievable classification accuracy and segmentation mIoU under both bpfp and bpp. Overall, larger backbones tend to provide stronger coded features, but the comparison between Base and Large reveals an important metric effect. In bpfp curves, DINOv2-L-Reg4 appears better than DINOv2-B-Reg4; however, in bpp-based BD metrics, DINOv2-B-Reg4 achieves better coding efficiency on ADE20K. This is because bpfp normalizes by feature points and can understate the cost of higher-dimensional features, whereas bpp is normalized by original image pixels. We therefore recommend bpp for cross-model comparison.

Table~\ref{tab:scale} reports BD metrics for DINOv2-Reg4 models, using DINOv2-S-Reg4 as the anchor.  On ImageNet, the rate--accuracy curves are nearly horizontal in the evaluated range, so we report BD-Acc only. These results indicate that, in distributed deployment scenarios, the largest backbone is not necessarily the best choice for feature coding. Instead, the backbone scale should be selected according to the target task, rate metric, and deployment objective.

\begin{table}[t]
    \centering
    \caption{BD metrics of VTC with different DINOv2-Reg4 backbone scales, using DINOv2-S-Reg4 as the anchor. ImageNet BD-rate is omitted because the rate--accuracy curves have insufficient overlap.}
    \label{tab:scale}
    \begin{tabular}{@{}lrrrr@{}}
        \toprule
        VTC        & \multicolumn{2}{c}{ADE20K val} & \multicolumn{2}{c}{ImageNet sel2k}                    \\
                   & BD-rate                        & BD-mIoU                            & BD-rate & BD-Acc \\
        \midrule
        DINOv2-S-Reg4 & 0.0\%                       & 0.00                               & -       & 0.00 \\
        DINOv2-B-Reg4 & -57.20\%                    & 3.57                               & -       & 4.62 \\
        DINOv2-L-Reg4 & -28.41 \%                   & 2.31                                 & --      & 7.47 \\
        \bottomrule
    \end{tabular}
\end{table}

\subsection{Ablation Study and Analysis}
\label{subsec:ablation}

The following studies isolate codec design choices on DINOv2-B-Reg4 under the multi-task-compatible ADE20K Seg protocol unless noted.
Table~\ref{tab:ablation} is introduced in Sec.~\ref{subsec:intermediate_layer}; here we focus on rows~4--5 and additional layout/latency comparisons.

\subsubsection{Effect of ViT Blocks in the Codec}
As shown in Table~\ref{tab:ablation} (rows~3--5), we analyze ViT blocks inside $f_a$ and $f_s$ at Input$=$9.
Removing blocks weakens the nonlinear analysis/synthesis transforms that decorrelate tokens before entropy coding.
One block costs 2.49 BD-mIoU; removing all blocks collapses to a near-linear pipeline and $-$15.51 BD-mIoU, indicating that feature coding needs capacity beyond a pure entropy model on raw tokens.

\subsubsection{Necessity of Dual-Path Design}
\label{subsubsec:dual_path_ablation}

\begin{figure}[t]
    \centering
    \includegraphics[width=\linewidth]{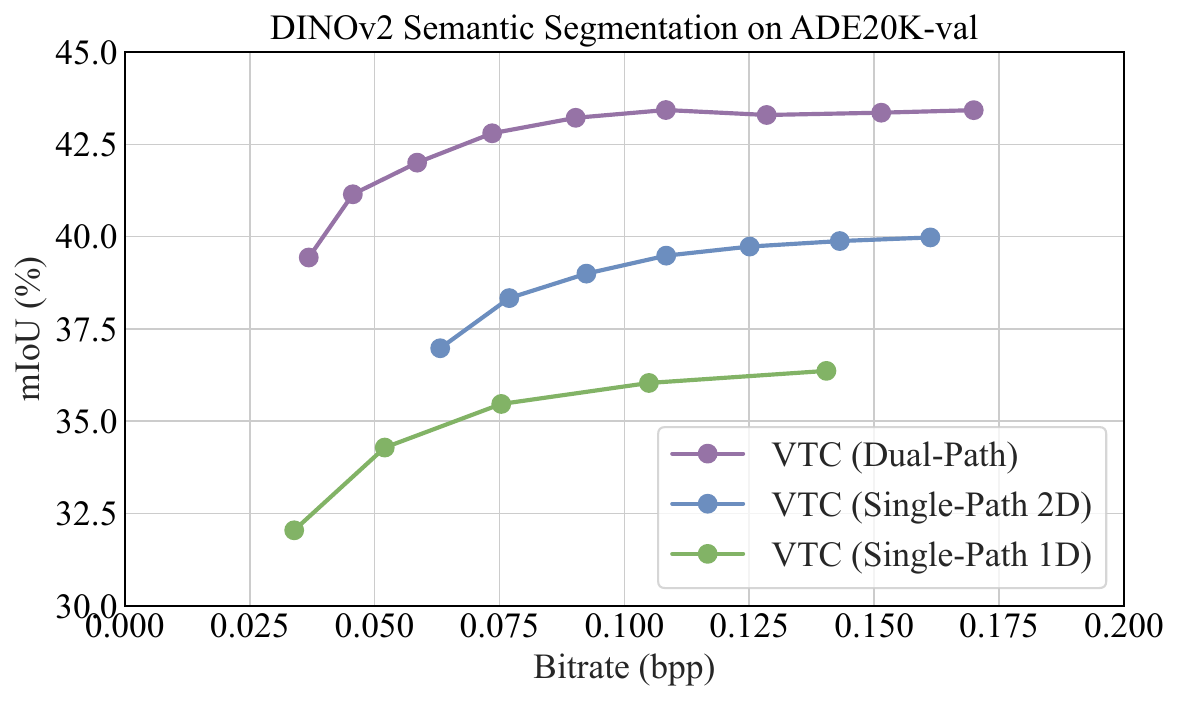}
    \caption{Rate--performance comparison of dual-path vs.\ all-1D or all-2D token layouts for entropy modeling.}
    \label{fig:dual_path_ablation}
\end{figure}

To verify that separate 1D/2D entropy models are needed (rather than a single layout for all tokens), we fix the ViT transform blocks and vary only token arrangement:
\textbf{(i) All-1D}: flatten all tokens into one sequence and compress with a factorized-prior model;
\textbf{(ii) All-2D}: arrange tokens with zero-padded global/register tokens in the top row followed by patch tokens, then compress with SCCTX;
\textbf{(iii) Dual-path} (ours): global tokens via factorized prior, patch tokens via SCCTX.
In Fig.~\ref{fig:dual_path_ablation}, all-1D performs worst because it ignores patch spatial structure; all-2D improves but padded global/register tokens pollute the SCCTX neighborhoods.
Dual-path keeps globals on a factorized 1D path and patches on SCCTX, matching token statistics and yielding the best RD curve.

\subsubsection{Wall-Clock Coding Latency}
\label{subsubsec:latency}

Table~\ref{tab:latency} reports end-to-end encoding/decoding time (seconds per image) on an RTX~4090 under the variable resolution setting.
LaMoFC-VTM relies on VTM, a traditional codec designed for video coding, and becomes computationally expensive when applied to high-dimensional feature tensors, limiting its practicality for deployment.
VQFC is much faster, requiring only 0.018\,s, but typically operates at around 4~bpp, which is substantially higher than JPEG image coding ($\sim$0.5~bpp).
Such a feature storage cost, exceeding that of storing the original images, is undesirable for large-scale deployment.
In contrast, VTC preserves high downstream task performance at $\sim$0.1~bpp with moderate encoding and decoding complexity, making it suitable for storage-oriented scenarios on large datasets.

\begin{table}[t]
    \centering
    \caption{Wall-clock coding time in seconds per image under the aligned setting. Runtime is measured on a single NVIDIA RTX~4090 GPU with an AMD EPYC 7Y43 CPU host.}
    \label{tab:latency}
\begin{tabular}{@{}lrrr@{}}
\toprule
Method  & LaMoFC VTM & VQFC  & VTC   \\ \midrule
Enc Time (s) & 580.884    & 0.018 & 0.091 \\
Dec Time (s) & 0.464      & 0.018 & 0.046 \\ \bottomrule
\end{tabular}
\end{table}

\section{Conclusion}
\label{sec:conclusion}
We presented Visual Token Codec (VTC), a deployment-oriented framework for compressing intermediate Vision Transformer features in distributed inference and feature-storage scenarios. Instead of treating ViT tokens as a flattened pseudo image, VTC separates global and patch tokens, applies spatial--channel context modeling on the native patch-token grid, and uses feature-matching supervision to preserve downstream semantics after decoding. Experiments on DINOv2 and SAM3 show that VTC consistently improves rate--performance trade-offs over existing feature coding baselines across five downstream tasks, reducing the bitrate required to reach 90\% of uncompressed-feature performance by 15.7$\times$--37.4$\times$. The multi-task, intermediate-layer, and backbone-scale analyses further indicate that VTC can produce reusable bitstreams, support practical computation--bandwidth trade-offs, and enable variable-rate coding within a single model. Future work will focus on faster entropy decoding, more reconstruction-aware objectives, and broader validation across foundation models and deployment constraints.






\bibliographystyle{IEEEtran}
\bibliography{references}

\end{document}